\documentclass[letterpaper]{article}
\usepackage{aaai18}
\usepackage{times}
\usepackage{helvet}
\usepackage{courier}
\usepackage{graphicx}
\usepackage{units}
\usepackage{amsfonts}
\usepackage{amsmath}
\usepackage{amssymb}
\usepackage{amsthm}
\usepackage{color}
\usepackage[colorinlistoftodos, textwidth=1.5cm, shadow]{todonotes}
\usepackage{ragged2e}
\usepackage{longtable}
\usepackage{colortbl}
\usepackage{titling}
\usepackage{wrapfig}
\makeatletter
\newcommand{\@BIBLABEL}{\@emptybiblabel}
\newcommand{\@emptybiblabel}[1]{}
\makeatother
\usepackage[hidelinks]{hyperref}

\graphicspath{{./figures/}}

\newcommand{\citet}[1]{\citeauthor{#1}~\shortcite{#1}}
\newcommand{\citep}{\cite}

\DeclareMathOperator*{\argmax}{\arg\, max\, }

\newtheoremstyle{example}
{3pt}
{3pt}
{\addtolength{\leftskip}{9pt}}
{}
{\bf}
{.}
{ }
{\thmname{#1}\thmnumber{ #2}\thmnote{ (#3)}}

\theoremstyle{example}

\newtheorem{ex}{Example}

\begin{document}

\title{Rethinking System Health Management}

\author{
  Edward Balaban \\
    {\normalsize\itshape
    Intelligent Systems Division, NASA Ames Research Center, Moffett Field, CA 94035}\\
		\and
	 Stephen B. Johnson \\
		{\normalsize\itshape
   Dependable System Technologies, LLC} \\
		{\normalsize\itshape
   Jacobs ESSCA Group at NASA Marshall Space Flight Center} \\
		\and
	 Mykel J. Kochenderfer \\
		{\normalsize\itshape
		Department of Aeronautics and Astronautics, Stanford University, Stanford, CA 94305} \\
	}
\date{}
\maketitle
\begin{abstract}
\begin{quote}
Health management of complex dynamic systems has traditionally evolved separately from automated control, planning, and scheduling (generally referred to in the paper as decision making). A goal of Integrated System Health Management has been to enable coordination between system health management and decision making, although successful practical implementations have remained limited. This paper proposes that, rather than being treated as connected, yet distinct entities, system health management and decision making should be unified in their formulations. Enabled by advances in modeling and computing, we argue that the unified approach will increase a system's operational effectiveness and may also lead to a lower overall system complexity. We overview the prevalent system health management methodology and illustrate its limitations through numerical examples. We then describe the proposed unification approach and show how it accommodates the typical system health management concepts.
\end{quote}
\end{abstract}

\section{Introduction}

System Health Management (SHM) has evolved from simple red-line alarms and human-initiated responses to a discipline that often includes sophisticated modeling and automated fault recovery recommendations \cite{Aaseng2001abr}. The main goal of modern SHM has been defined as the preservation of a system's ability to function as intended \cite{Rasmussen2008,Johnson2011a}. While in this paper we apply the term SHM to the operational phase of a system's lifespan, in other contexts the term may also encompass design-time considerations \cite{Johnson2011a}. The actual achievement of a system's operational objectives, on the other hand, is under the purview of the fields of control, planning, and scheduling. In this paper, we will generally refer to all of the processes aimed at accomplishing operational objectives as decision making (DM), while using the more specialized terms where necessary.

Historically, the field of SHM has developed separately from DM. The typical SHM functions (monitoring, fault detection, diagnosis, mitigation, and recovery) were originally handled by human operators--and similarly for DM \cite{Aaseng2001abr,Ogata2010}. As simple automated control was being introduced for DM, automated fault monitors and alarms reduced operator workload on the SHM side. Gradually, more sophisticated control techniques were developed for DM \cite{Ogata2010}, while automated emergency responses started handling some of the off-nominal system health conditions \cite{Rasmussen2008}. Capable automated planning and scheduling tools eventually became available for performing more strategic DM \cite{Ghallab2016}. On the SHM side, automated fault diagnosis was added, in some cases coupled with failure prediction (i.e., \textit{prognostic}) algorithms \cite{Lu1979}, as well as with recovery procedures \cite{Avizienis1976}. Still, the two sides largely remained separated. When the concept of Integrated System Health Management became formalized, most interpretations of \textit{integrated} encompassed some interactions between DM and SHM, although practical implementations of such interactions have been limited \cite{Figueroa2018}.

This paper makes the following claims:

\begin{enumerate}

	\item For actively controlled systems, prognostics is not meaningful; for uncontrolled systems prognostics may only be meaningful under a specific set of conditions;
	\item DM should be unified with SHM for optimality and there is a path for doing so;
	\item Automated emergency response should be done separately from unified DM/SHM, in order to provide performance guarantees and dissimilar redundancy.

\end{enumerate}

For the second claim, we intend to show a path towards DM/SHM unification that builds on the latest developments in state space modeling, planning, and control. While in the past limitations in computing hardware and algorithms would have made unified DM/SHM difficult, we believe that the advances of the recent years make it an attainable goal. We also believe that this unified approach is applicable to a broad spectrum of DM, from traditional deterministic planners to complex algorithms computing long-horizon action policies in the presence of uncertainty. We use the term \emph{unification} to emphasize the idea of DM and SHM being done within the same framework, rather than the two being integrated as separate subsystems exchanging information.

Some initial progress towards DM/SHM unification can be found in the earlier work by \citet{Balaban2013a}, \citet{Balaban2018}, and \citeauthor{Johnson2010} (\citeyear{Johnson2010,Johnson2011a}). System health information was also incorporated into DM by others \cite{Bethke2008,Ure2013,Agha-mohammadi2014}. This paper aims to introduce a systematic view on such integration, discuss its benefits, and illustrate how current SHM concepts map into the proposed approach without a loss of functionality or generality.

The paper first defines the categories of systems that are of interest to this study (Section \ref{sec:systems}) and then overviews the prevailing approach to  SHM (Section \ref{sec:shm}). The first claim (on prognostics) is discussed in Section \ref{sec:prognostics}. Section \ref{sec:hadm} discusses the second claim, concerning DM/SHM integration and its benefits, as well as the rationale for the third claim (that DM/SHM should be separate from automated emergency response). Section \ref{sec:conclusions} concludes.

\section{Systems of interest}
\label{sec:systems}

In the discussion to follow, we consider both uncontrolled and controlled systems. For our purposes, the \textbf{uncontrolled systems} category includes not only those systems types for which control is not available or required, but also those operating on predefined control sequences, such as industrial robots performing the same sets of operations over extended periods of time.  Also included are system types that can be considered uncontrolled within some time interval of interest (\textbf{decision horizon}). Systems in this category may have degradation modes that affect the system's performance within its expected useful life span. The rate of degradation is influenced by internal (e.g., chemical decomposition) and external factors (e.g., temperature of the operating environment). In addition to aforementioned industrial robots, examples of uncontrolled system types include bridges, buildings, electronic components, and certain types of rotating machinery, such as electrical power generators.

The \textbf{controlled systems} category covers all other system types, including dynamically controlled systems operating in uncertain environments, where the current state cannot be fully observed and non-determinism is present in control action outcomes. Degradation processes are influenced not only by the same kinds of internal and external factors as for the uncontrolled systems, but also by the control actions.

Most of the discussion to follow is applicable to both categories, although controlled systems would, naturally, benefit more from active DM/SHM. In describing the systems, we adopt the notation from the field of decision making under uncertainty \cite{Kaelbling1998}. 

A system \textbf{state} $s$ can be a scalar or a vector belonging to a state space $S$. A system \textbf{action} $a$ initiates state transitions, with $A$ denoting the space of all available actions ($A$ may be state-dependent). A \textbf{transition model} $T(s,a,s')$ describes the probability of transitioning to a particular state $s' \in S$ as a result of taking action $a$ from state $s$. A \textbf{reward model} $R(s,a)$ describes a positive reward obtained or a negative cost incurred as a result of taking action $a$ from state $s$. \textbf{Terminal states} form a subset $S_T \subset S$. Terminal states $S_T$ may include both failure and goal states. Transitions from a terminal state are only allowed back to itself.

If there is state uncertainty, \textbf{belief states} are used instead of regular states (also referred to as \textbf{beliefs}). A belief $b$ is a probability distribution over $S$, with $B$ denoting the space of all beliefs. \textbf{Observations} (\textit{e.g.}, sensor readings) can help with \textbf{belief estimation} and \textbf{updating}. Like a state, an observation can be a vector quantity. An \textbf{observation model} $O(s',a,o)$ describes the probability of getting an observation $o$ upon transition to state $s'$ as a result of action $a$.

The general function of decision making is to select actions. While in some systems we are only concerned with selecting a single $a_t$ at a given time $t$, decision making problems often involve selecting a sequence of actions. In a strictly deterministic system, an entire sequence of actions (a \textbf{plan}) can be selected ahead of time. In systems with action outcome uncertainty, however, a fixed plan can quickly become obsolete. Instead, a \textbf{policy} $\pi(s): S \rightarrow A$ needs to be selected that prescribes which action should be taken in any state. If a policy is selected that, in expectation, optimizes a desired metric (\textit{e.g.}, maximizes cumulative reward), it is referred to as an \textbf{optimal policy} and denoted $\pi^*$.

Throughout the paper, we use a robotic exploration rover operating on the surface of the Moon as a running example of a complex controlled system. The rover is solar-powered and stores electrical energy in a rechargeable battery.

\section{System Health Management}
\label{sec:shm}

\begin{figure}[ht]
	\centering
		\includegraphics[width=\columnwidth]{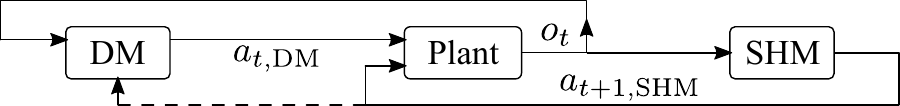}
	\caption{A typical system architecture with SHM}
	\label{fig:system}
\end{figure}

A typical contemporary SHM integration approach is shown in Figure \ref{fig:system}. A DM subsystem generates an action $a_{t,\text{DM}}$, aimed at achieving the operational objectives. The plant executes $a_{t,\text{DM}}$ and an observation $o_t$ is generated and relayed to both DM and SHM. DM computes $a_{t+1,\text{DM}}$ on the basis of $o_t$, while SHM analyzes $o_t$ for indications of \textbf{faults} (defined here as system states considered to be off-nominal) and, if any are detected, issues a mitigation or recovery command $a_{t+1,\text{SHM}}$ either directly to the plant or as a recommendation to the DM subsystem \cite{Valasek2012}.

\begin{figure*}[ht!]
	\centering
		\includegraphics[width=0.9\textwidth]{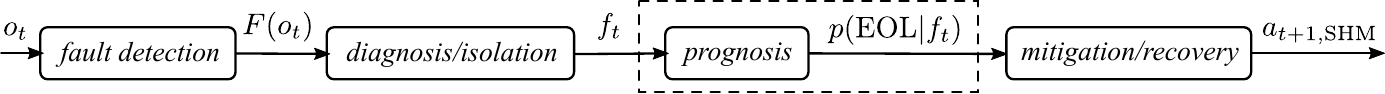}
	\caption{A typical contemporary SHM architecture}
	\label{fig:shm}
\end{figure*}

Figure \ref{fig:shm} goes into more detail on the SHM subsystem. The fault detection module corresponds to the traditional red-line monitors detecting threshold-crossing events of sensor values, represented on the diagram by a Boolean \textbf{fault detection} function $F$. If a fault is detected ($F(o_t) = \text{true}$), \textbf{fault isolation} and \textbf{diagnosis} (or \textbf{identification}) is performed, generating a vector of fault descriptors $f_t$ \cite{Daigle2010}. Each fault descriptor typically identifies a component, its fault mode, and its fault parameters \cite{Daigle2010}. There are diagnostic systems that also include an estimated fault probability in the descriptor \cite{Narasimhan2007}. If the uncertainty of the diagnostic results is deemed too high (\textit{e.g.}, $f_t$ consists of only low-probability elements), \textbf{uncertainty management} is sometimes performed in order to obtain a better estimate of the current system condition \cite{Lopez2010}.

Some recent SHM implementations then pass $f_t$ to a \textbf{prognostic} module \cite{Roychoudhury2011}. In the SHM context, the intended goal of the prognostic module is to predict, at time $t_p$ (here $t_p=t$), whether and when faults will lead to system \textbf{failure} (defined as inability to perform its assigned function) within the window $[t_p,t_p+H]$ of a prediction horizon $H$ (the terms \textit{prediction horizon} and \textit{decision horizon} are equivalent for our purposes). In prognostics literature, the time of failure is commonly synonymous with the term \textbf{end of [useful] life (EOL)}. Equivalently, the goal of prognostics can be defined as predicting the system's \textbf{remaining useful life (RUL)}. In Figure \ref{fig:shm}, the prognostic prediction is represented  as a probability distribution $p(\text{EOL}|f_t)$ for $\text{EOL} \in [t_p,t_p+H]$. Uncertainty management is sometimes also prescribed following prognostic analysis, meant to improve the prediction if the confidence in it is insufficiently high \cite{Wang2012}. Note that if a prognostic module is part of an SHM sequence, the term \textbf{Prognostics and Health Management (PHM)} is used by some instead of SHM in order to emphasize the role prognostics is playing in managing the system's lifecycle.

Finally, $p(\text{EOL}|f_t)$ and $f_t$ are passed to the \textbf{fault mitigation and recovery} component to select an action $a_{t+1,\text{SHM}}$ from the action set $A_{\text{SHM}}$, in order to mitigate or recover from faults in $f_t$. As part of this process, operational constraints may be set for those faulty components that cannot be restored to nominal health. If functional redundancy exists for such components, their further use may be avoided.

The overall limitations of the current SHM approach are discussed in Section \ref{sec:hadm}, where an approach that unifies DM and SHM is then proposed. The next section, however, focuses on the prognostic component and discusses why it is not meaningful for actively controlled systems and is challenging to implement in a useful manner for uncontrolled systems.

\section{Prognostics}
\label{sec:prognostics}

A general definition of prognostics is that of a process predicting the time of occurrence of an event $E$ \cite{Daigle2015}. Using notation from Section \ref{sec:systems}, if $\phi_E: S \rightarrow \mathbb{B}$ (where $\mathbb{B} \triangleq \{0,1\}$) is an event threshold function, then $t_E \triangleq \text{inf}\{t \in [t_p,t_p+H]: \phi_E(s_t) = 1\}$ is the nearest predicted time of $E$ ($s_t$ is the state at time $t$). If the state evolution trajectory is non-deterministic, then $p(t_E|s_{0:t_p})$ is computed instead. If states cannot be directly observed, $p(t_E|o_{0:t_p})$ is computed. As defined, prognostics is only meaningful in a specific set of circumstances and next we use two examples to illustrate why.

\subsection{Uncontrolled systems}

There are two main desirable, interrelated attributes for a prognostic system: (1) low uncertainty in EOL estimation, so that a decision about mitigation or recovery actions can be made with confidence, and (2) the ability to make a prediction far enough in advance for the actions to be successfully executed. For uncontrolled systems, this means that prognostics is primarily useful for systems with long lifetimes, low process uncertainty, or both. To illustrate why this is the case, we start with a simple uncontrolled system example:

\begin{figure}[h]
	\centering
		\includegraphics[width=0.9\columnwidth]{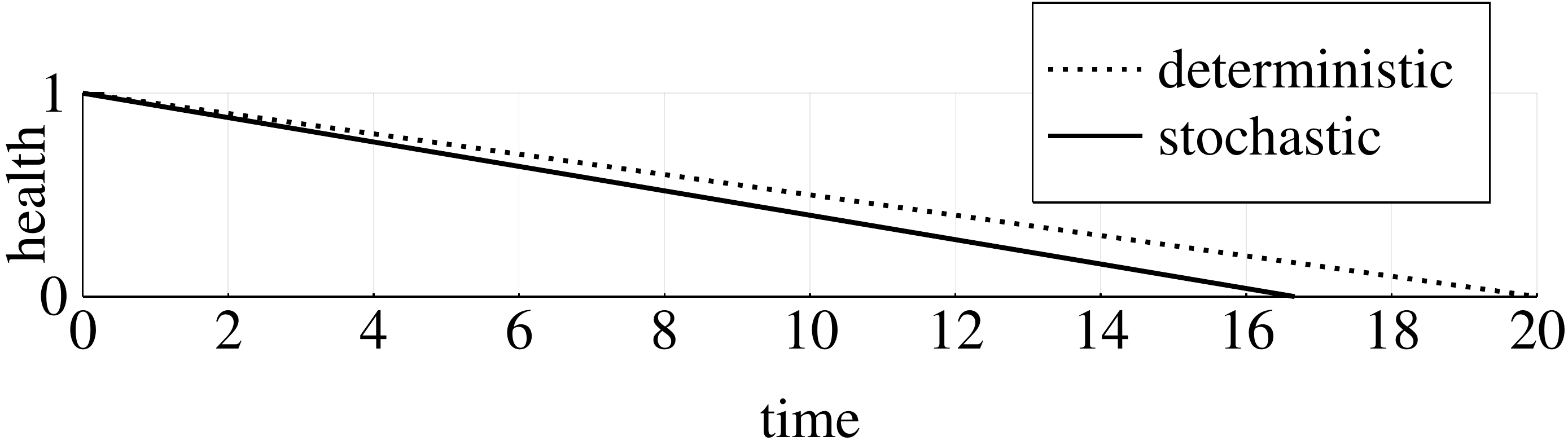}
	\caption{Uncontrolled system prognostics (Example \ref{ex:eol})}
	\label{fig:prognostics}
\end{figure}

\begin{ex}
\label{ex:eol}

At time $t=0$, the health state of a system is $s_0 = 1$ (states are scalar). According to a deterministic model, the nominal health degradation rate is constant at $\dot s_n = 0.05/\Delta t$, where $\Delta t$ is the prediction time step, selected as the minimum time interval within which a change in the system's health is expected to be detectable. A stochastic model predicts the probability of the nominal degradation rate $\dot s_n$ within any time step as $p_n = 0.8$ and the probability of a higher degradation rate ($\dot s_h= \dot s_n + \epsilon / \Delta t$) as $p_h = 0.2$. Assume $\epsilon = 0.05$.

The objective for both models is to predict EOL, \textit{i.e.}, the smallest $t$ for which $s \leq 0$. For this example, the prediction uncertainty is defined as $\sigma(t_p) = |E[\text{EOL}_d(t_p)] - E[\text{EOL}_s(t_p)]|$, \textit{i.e.}, the absolute difference between the expected EOL values computed by the two models at a prediction time $t_p$. A requirement is set on the maximum EOL prediction uncertainty as $\sigma_\textit{max} = 1 \Delta t$. 
\end{ex}

For this example, let us assume that the health state is fully observable and define $\rho_p = s_p/s_0$, the fraction of full health remaining at $t_p$. In Figure \ref{fig:prognostics}, a prediction is shown to be made at $t_p = t_0$, with $s_p = s_0$, $\rho_p = 1$, and $H=20 \Delta t$. Since

\begin{align*}
\begin{split}
	& E[\text{EOL}_d(t_p)] = \dfrac{\rho_p}{\dot s_n} \text{ and } E[\text{EOL}_s(t_p)] = \dfrac{\rho_p}{p_n\dot s_n + p_h \dot s_h} = \\
	& = \dfrac{\rho_p}{(1-p_h)\dot s_n + p_h (\dot s_n+\epsilon / \Delta t)} = \dfrac{\rho_p}{(\dot s_n + p_h \epsilon / \Delta t))}, \text{ then}\\
\end{split}
\end{align*}
\begin{equation}
\sigma = \left|\dfrac{\rho_p}{\dot s_n} - \dfrac{\rho_p}{(\dot s_n + p_h \epsilon / \Delta t)} \right| = \rho_p \left |20 - \dfrac{1}{(0.05 + p_h \epsilon)} \right|\Delta t.
\label{eq:sigma}
\end{equation}

In the last equation, we substitute the value for the nominal degradation rate $\dot s_n$ in order to focus on the effects of the degradation rate uncertainty. As can be seen in Figure \ref{fig:prognostics}, EOL is reached by both models within the prediction horizon. However, from Equation \ref{eq:sigma}, $\sigma  = 3.33\Delta t > \sigma_{max}$.

For the requirement on $\sigma$ to be satisfied, either $\rho_p$ (health fraction remaining), $p_h$ (the probability of deviations from the nominal degradation rate), $\epsilon$ (the magnitude of deviations), or some combination of them needs to be reduced. If $p_h$ and $\epsilon$ are kept the same, with $\rho_p = 0.25$ we can get $\sigma  = 0.83\Delta t$. However, $t_p$ now needs to be $\approx 15 \Delta t$, with only $5 \Delta t$ left until failure (RUL). RUL corresponds to the time available to either replace/repair the uncontrolled system or initiate an emergency response. For a quickly degrading system with $\Delta t = \unit[1]{s}$, $\text{RUL}$ would be only $\unit[5]{s}$, which is likely enough time for an emergency response, but not for repair or replacement. In practice, in uncontrolled systems where handling a fast-developing degradation process is important, estimating $p(\text{EOL})$ is unlikely to bring tangible benefits. For instance, if pressure starts building quickly in a fuel tank of an ascending crewed rocket, the launch abort system (emergency response) is likely to be activated by the exceedance of a predefined pressure limit or pressure increase rate (\textit{i.e.}, functions of fault detection). Computing whether the tank breach will occur in $10$ seconds or $12$ seconds will not materially influence that response. For uncontrolled systems with mid-range degradation rates (minutes, hours, days), extrapolation of the degradation function may be able to serve as part of a caution and warning mechanism.

What follows is that if degradation uncertainty is relatively high or varies significantly over time, a short prediction horizon (compared to the overall system lifetime) may be necessary to limit the uncertainty propagation and result in a usable $\sigma$. In this case, systems with longer lifetimes are more suitable for applying prognostics. For example, if a bridge failure can be predicted $3$ years in advance with an accuracy of $\pm 1$ year, that can still be a useful prediction.

However, while many uncontrolled systems can be classified as systems with long lifetimes, there exists a number of fundamental practical difficulties in performing effective prognostics for them. One of the primary issues stems directly from the typically long (often decades) lifetimes. In order to establish trust in the degradation models, they need to be adequately tested using long-term observations from ``normal use'' degradation or observations from properly formulated accelerated testing. With useful (from the statistical point of view) ``normal use'' degradation data sets being rare for many long-life system types \cite{Heng2009}, accelerated degradation efforts are common. If an accelerated degradation regime is proposed, however, what needs to be clearly demonstrated is that:

\begin{enumerate}
	\item \textbf{The regime can be used as a substitute for real-life degradation processes}. For instance, while \citeauthor{Rigamonti2016} \shortcite{Rigamonti2016} and \citeauthor{Celaya2011a} \shortcite{Celaya2011a} use thermal and electrical overstress to quickly degrade electrical capacitors and predict the time of their failure (by using an empirical equation), their work does not extend to making a connection to real-life degradation, which takes place at lower temperatures and voltage/current levels. Similar issues are highlighted by \citeauthor{Dao2006} \shortcite{Dao2006} for composite materials, where mechanical, thermal, and chemical processes result in complex interactions during aging.
	
	\item \textbf{There is a known mapping from the accelerated timeline to the unaccelerated timeline.} \citeauthor{Oh2015} \shortcite{Oh2015} note in an overview of condition monitoring and prognostics of insulated gate bipolar transistors that while numerous fatigue models have been constructed that predict \textit{cycles-to-failure} under repetitive cycle loading, they are not designed to predict RUL under realistic usage conditions. Although use of various fatigue analysis models, \textit{e.g.}, \citeauthor{Paris1961} \shortcite{Paris1961}, has been proposed for estimating RUL on the basis of stress cycles, their accuracy has proven difficult to confirm.
\end{enumerate}

There are subfields of prognostics where accelerated aging regimes may be viable, such as in aircraft metallic structures or rotating machinery (where mechanical degradation factors could be assumed dominant). However, the issue of high uncertainty of degradation trajectories still arises, even under the same test conditions \cite{Virkler1979,Meng1994}. Finite element modeling may help alleviate degradation trajectory uncertainty in specific cases, albeit at a significant computational cost \cite{Heng2009}.

Some of the other challenges with effective prognostics for uncontrolled systems include the accuracy of estimating the actual state of health, the effects of fault interactions, and the effects of system maintenance actions \cite{Heng2009}.

If these challenges are successfully overcome and the failure mechanisms of a component are understood well enough to develop useful degradation models, a different question then arises: should the design or usage of the component be changed to mitigate these mechanisms?  While in some cases this may not be feasible, in others it may be the simplest and most reliable way of improving safety and maintenance requirements \cite{Bathias2013}. A redesign or change in usage would, on the other hand, make the degradation models obsolete. The next tier of degradation modes would then need to be analyzed and modeled, possibly followed by another redesign. Thus analysis intended for the development of degradation (prognostics) models instead becomes part of the design cycle.

For those uncontrolled systems that are, in fact, suitable for health management based on prognostics, the action space is typically limited to: (a) no action, (b) replacement, or (c) repair to a nominal operating condition. Even so, we still propose that predictive analysis for these systems needs to be driven by decision making requirements. For instance, if domain knowledge informs that variability in system behavior beyond some health index value $h_\text{min}$ is too great for choosing actions with sufficiently high confidence, then EOL can be redefined as $h_\text{min}$ and system dynamics beyond $h_\text{min}$ need to be neither modeled nor computed during prediction, potentially freeing computing resources for estimating system behavior up to $h_\text{min}$ with more accuracy.

\subsection{Controlled systems}

\begin{figure}[h!]
	\centering
		\includegraphics[width=0.82\columnwidth]{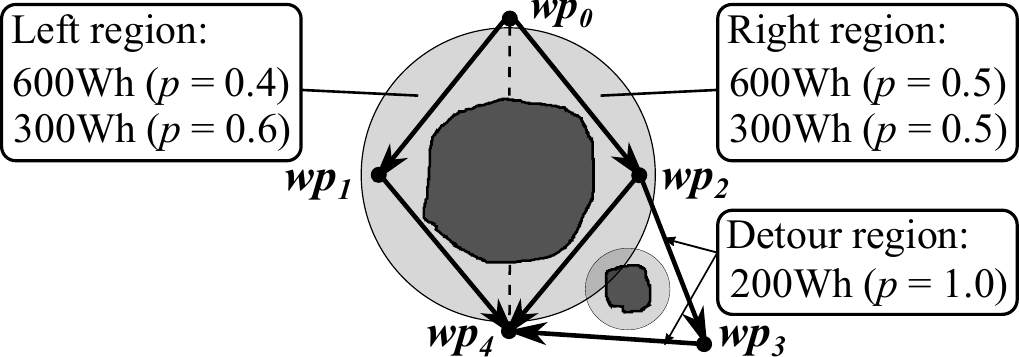}
	\caption{PHM vs. DM for a controlled system (Example \ref{ex:phm_dm})}
	\label{fig:phm_dm_example}
\end{figure}

As soon as dynamic control is introduced into the system and uncertainty is taken into account, prognostics, as defined above, and the PHM version of the process in Figure \ref{fig:shm} are no longer meaningful--for two key reasons. First, not having the knowledge, at $t_p$, of the future system actions, a PHM algorithm will either (a) have to rely on some precomputed plan to obtain $a_{t_p+1:H}$ for its predictive analysis (a plan that can quickly become obsolete due to outcome uncertainty) or (b) have to use a random policy (which can, for instance, result in less optimal actions being considered as probable as the more optimal ones). A random policy is also likely to result in a greater state uncertainty throughout the $[t_p,t_p+H]$ interval. Second, the oft-proposed strategy of rerunning prognostic analysis after an action, so that new information can be taken into account \cite{Tang2011}, may not help. Once a suboptimal execution branch has been committed to, it may remain suboptimal regardless of future decisions. The following example provides an illustration of these issues:

\begin{ex}
\label{ex:phm_dm}
A rover needs to traverse an area with no sunlight, going around a large crater from waypoint $wp_0$ to the closest suitable recharge location at $wp_4$. The battery charge at $wp_0$ is $\unit[1100]{Wh}$.

There are three possible levels of terrain difficulty: \textit{difficult} (requiring $\unit[600]{Wh}$ per drive segment), \textit{moderate} ($\unit[300]{Wh}$ per segment), and \textit{easy} ($\unit[200]{Wh}$ per segment). All drive segments are the same length. Probabilities of terrain types in different regions are shown in Figure \ref{fig:phm_dm_example}.

The rover can go to the left, $wp_0 \rightarrow wp_1 \rightarrow wp_4$, or to the right, $wp_0 \rightarrow wp_2 \rightarrow wp_4$ (left and right are relative to the diagram). If going to the right, it can decide to detour around a smaller crater $wp_2 \rightarrow wp_3 \rightarrow wp_4$ (\textit{easy} terrain with $p=1.0$) instead of going directly $wp_2 \rightarrow wp_4$.

\end{ex}

\begin{figure}[h!]
	\centering
		\includegraphics[width=0.9\columnwidth]{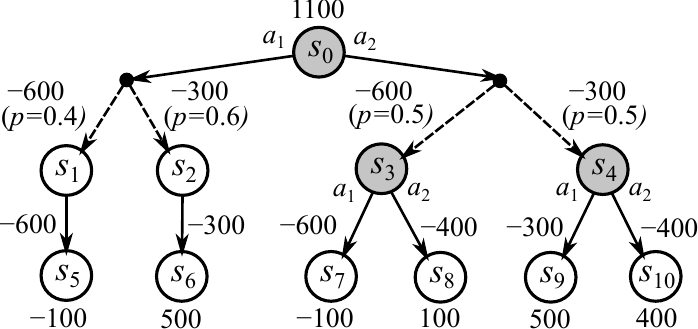}
	\caption{Execution scenarios in Example \ref{ex:phm_dm}}
	\label{fig:phm_dm_tree}
\end{figure}

A decision support PHM algorithm, running a sufficiently large number of simulations, would consider two possible execution scenarios along the left route: (L1) $e_\text{total}=\unit[1200]{Wh}$, $p=0.4$ and (L2) $e_\text{total}=\unit[600]{Wh}$, $p=0.6$ ($e_\text{total}$ is the total energy consumed in a scenario). The expected energy consumption along the left route can then be computed as $E[e_\text{total, L}]=\unit[1200]{Wh} \cdot 0.4 + \unit[600]{Wh} \cdot 0.6 = \unit[840]{Wh}$.

The algorithm would then consider four possible execution scenarios along the right route (assuming uniform random choice of action at $wp_3$): (R1) $e_\text{total}=\unit[1200]{Wh}$, $p=0.25$; (R2) $e_\text{total}=\unit[600]{Wh}$, $p=0.25$; (R3) $e_\text{total}=\unit[1000]{Wh}$, $p=0.25$; and (R4) $e_\text{total}=\unit[700]{Wh}$, $p=0.25$. Then $E[e_\text{total, R}]=\unit[(1200+600+1000+700) \cdot 0.25]{Wh} = \unit[875]{Wh}$. With $E[e_\text{total, L}] < E[e_\text{total, R}]$, the PHM algorithm commits to the left path. Note that in many cases prognostics algorithms generate even less information to support action selection than what was done here (typically $p(\text{EOL})$ only).

A DM algorithm capable of sequential reasoning under uncertainty \cite{Kochenderfer2015} would compute $E[e_\text{total, L}] = \unit[840]{Wh}$ in the same manner as the PHM algorithm, as there are no actions needing to be selected along the left route after $wp_0$. On the right side, however, the DM algorithm can make an informed choice at $wp_2$, based on observations made along $wp_0 \rightarrow wp_2$. This implies having only two possible execution scenarios: (R1) if the terrain is observed as \textit{difficult}, the detour through $wp_3$ is taken, and (R2) if the terrain is observed as \textit{moderate}, the rover goes directly to $wp_4$. For R1, $e_\text{total}=\unit[1000]{Wh}$, $p=0.5$. For R2, $e_\text{total}=\unit[600]{Wh}$, $p=0.5$. The expected energy use is thus $E[e_\text{total, R}]=\unit[(1000+600) \cdot 0.5]{Wh} = \unit[800]{Wh}$. With $E[e_\text{total, L}] > E[e_\text{total, R}]$, the algorithm chooses the right path.

Now let us assume that the true terrain condition both on the left and the right sides of the crater is \textit{difficult}. The left path ($wp_0 \rightarrow wp_1 \rightarrow wp_4$) will require $\unit[1200]{Wh}$ to traverse, therefore a rover relying on the PHM algorithm will fall $\unit[100]{Wh}$ short and will not reach $wp_4$. A rover relying on the DM algorithm will expend only $\unit[1000]{Wh}$ (scenario R1), arriving at $wp_4$ with $\unit[100]{Wh}$ in reserve.

It may be suggested that the issues with the PHM approach could be eliminated if access to a precomputed operational policy $\pi_\text{DM}$ is provided. However, even if such a policy was available, that would still be insufficient. If, at time $t$, $p(\text{EOL})$ is computed using $\pi_\text{DM}$, then $a_{t+1,\text{SHM}}$ is taken on the basis of $p(\text{EOL})$, $p(\text{EOL})$ could immediately become invalid unless $T(s_t,a_{t+1,\text{SHM}}, s_{t+1}) = T(s_t,a_{t+1,\text{DM}}, s_{t+1})$.

\section{Unifying DM and SHM}
\label{sec:hadm}

This section discusses the benefits of a unified DM/SHM approach and outlines some of the key implementation details.

\subsection{Separated vs. unified DM/SHM}

The next example illustrates how unification can be helpful in balancing system health needs and operational objectives:

\begin{figure}[h]
	\centering
		\includegraphics[width=0.85\columnwidth]{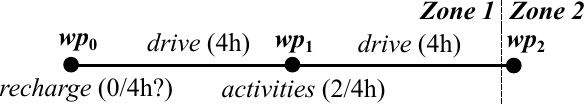}
	\caption{Separated vs. unified DM/SHM (Example \ref{ex:separated_unified})}
	\label{fig:shm_dm_example}
\end{figure}

\begin{figure}[h]
	\centering
		\includegraphics[width=0.75\columnwidth]{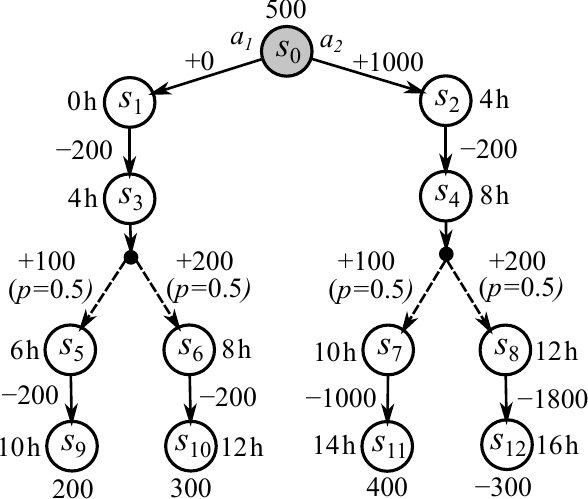}
	\caption{Execution scenarios in Example \ref{ex:separated_unified}}
	\label{fig:shm_dm_tree}
\end{figure}

\begin{ex}
\label{ex:separated_unified}
The rover starts at $wp_0$ in Zone 1 (Figure \ref{fig:shm_dm_example}) with $\unit[500]{Wh}$ in the battery (out of the $\unit[1500]{Wh}$ capacity). The solar panels can charge the battery at a rate of $\unit[250]{W}$. Sunlight in Zone 1 will last for another 12 hours.

Two actions are available in the initial state $s_0$ at $wp_0$ (Figure \ref{fig:shm_dm_tree}): skip charging ($a_1 = + \unit[0]{Wh}$) and charge to full ($a_2 = + \unit[1000]{Wh}$). In Figure \ref{fig:shm_dm_tree}, the unitless numbers are Watt-hours of energy going in or out of the battery. Time (in hours) at each state is denoted as '$\unit[\text{[t]}]{\text{h}}$'.

The rover needs to perform a 2-hour stationary science activity at $wp_1$ and be able to arrive at $wp_2$, the next recharge point. The prior probability of the activity at $wp_1$ needing to be redone (a 2-hour delay) is $0.5$. Science payload power consumption is $\unit[100]{W}$, resulting in a net-positive ($\unit[250]{W} -\unit[200]{W} = +\unit[50]{W}$) power flow.

The driving times from $wp_0$ to $wp_1$ and from $wp_1$ to $wp_2$ are $4$ hours, with the average drive train power consumption of $\unit[300]{W}$, resulting in a net-negative ($\unit[250]{W} -\unit[300]{W} = -\unit[50]{W}$) power flow.

If operating without sunlight, a $\unit[150]{W}$ heater needs to be used to keep batteries and electronics warm, thus resulting in a net-negative power flow of $-\unit[300]{W} -\unit[150]{W} = - \unit[450]{W}$ for driving and $- \unit[200]{W} - \unit[150]{W} = - \unit[350]{W}$ for stationary science activities.
\end{ex}

According to the general SHM policy of restoring health (battery charge, in this case) to nominal, the action chosen at $wp_0$ is $\pi_{shm}(s_0) = a_2$ and the battery is recharged to full $\unit[1500]{Wh}$. After the science activity at $w_1$ is completed, assessment is made that it needs to be repeated. The 2-hour delay means that the entirety of $wp_1 \rightarrow wp_2$ segment needs to be done without sunlight, resulting in a complete battery depletion before $wp_2$ is reached, with a deficit of $\unit[300]{Wh}$.

In computing a unified policy, however, where SHM actions are considered in the context of the overall mission, all four scenarios depicted in Figure \ref{fig:shm_dm_tree} would play a role. The expected amount of battery charge remaining if $a_1$ is chosen would be $Q_1 = 0.5 \cdot 200 + 0.5 \cdot 300 = 250$. For $a_2$: $Q_2 = 0.5 \cdot 400 + 0.5 \cdot (-300) = 50$. Action $a_1$ (no recharge) would be chosen and, with the two-hour delay at $wp_1$, the rover would arrive at $wp_2$ with $\unit[300]{Wh}$ still remaining.

The example illustrates the first benefit of unifying DM and SHM: the ability to naturally take operational objectives and constraints into account when making a system health recovery decision. The next example illustrates how DM, on the other hand, can benefit from unified action spaces and access to health-related models:

\begin{figure}[h]
	\centering
		\includegraphics[width=0.7\columnwidth]{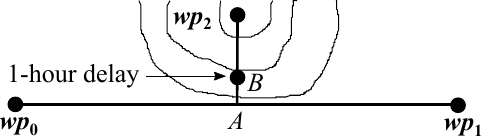}
	\caption{Unified action spaces (Example \ref{ex:action_spaces})}
	\label{fig:action_spaces}
\end{figure}

\begin{ex}
\label{ex:action_spaces}

The rover is traveling from $wp_0$ to $wp_1$ (flat terrain) when a decision is made at point A to make a detour and attempt data collection at a scientifically valuable $wp_2$, requiring a 6-hour climb up a steep hill (Figure \ref{fig:action_spaces}). The 1-hour data collection activity at $wp_2$ must be completed before the loss of illumination there in $10$ hours. 
	
After completing the 2-hour climb up to point $B$ ($1/3$ of the way up), it is observed that the internal temperature of one of the drive motors has risen to $T_m=\unit[60^\circ]{C}$ from the nominal of $\unit[20^\circ]{C}$. At $T_m=\unit[80^\circ]{C}$ there is a significant risk of permanent damage and failure of the motor.
\end{ex}

If the SHM system on the rover consists of the traditional fault detection, diagnosis, and mitigation/recovery components only, it may diagnose the fault to be \textit{increased friction}, mark the component (motor) as faulty, and set a constraint on terrain types to either \textit{flat} or \textit{downhill}. It would then invoke the recommended action for this fault from $A_\text{SHM}$: \textit{stop and cool down} (until $T_m=\unit[20^\circ]{C}$). 

If a prognostic component is present, it may predict that at the current temperature increase rate, the RUL of the motor is 1 hour (with 4 hours of climb remaining to reach $wp_2$). The same mitigation action (\textit{stop and cool down}) would be initiated and the same constraint on the current to the motor (and, thus, on the incline angle) may be set. After $T_m$ returns to nominal (which happens to take 1 hour), control is returned to DM. Based on the new constraints and the motor marked as faulty, DM would command the rover to abort the detour, return to $A$, and resume the drive to $wp_2$.

If, however, DM had \textit{stop and cool down} as part of its action space ($A_\text{DM}$) and updated the state variables for the affected motor with the newly computed heat-up and cool down rates, an operational policy could be computed that optimizes the duration of driving and cool down intervals and allows the rover to reach $wp_2$ in time. For instance, if the rover drives for two hours, then stops for an hour to cool down the motor, it can still reach $wp_2$ in $2+1+2+1+2 = 8$ hours. With the science activity taking 1 hour, there would still be an hour in reserve before the loss of sunlight at $wp_2$.

\subsection{Unification approach}
In proposing the unified DM/SHM approach, we rely heavily on \textit{utility theory} \cite{Fishburn1970}. The following, in our opinion, are the key ingredients for a successful  unification: (1) a state-based system modeling framework and (2) a utility (value) function capturing the operational preferences for the system. A \textbf{utility function} (denoted as $U$) captures, numerically, our preferences over the space of possible outcomes. For instance, we may assign a higher utility value to a rover state where a desired scientific location has been reached. Utility can be defined for system states or for state-action pairs. If, in some system state $s$, the outcome of an action $a$ (a particular state $s'$) is not guaranteed, then the \textit{expected utility} of the $(s,a)$ tuple is
\begin{equation}
U(s,a) = \sum_{s'}T(s',a,s)U(s').
\end{equation}

In a strictly deterministic system, where a plan $\{a_{0:H}\}$ up to a horizon $H$ can be provided ahead of time, the expected utility of $s$ relative to the plan is:
\begin{equation}
U^{a_{0:H}}(s) = \sum_{i=0:H}R(s_i,a_i).
\label{eq:plan}
\end{equation}
For systems with action outcome uncertainty, the expected utility associated with executing a policy $\pi$ for $t$ steps from state $s$ can be computed recursively as
\begin{equation}
U_t^{\pi}(s)=R(s,\pi(s))+\gamma \sum_{s'} T(s,\pi(s),s')U^{\pi}_{t-1}(s'),
\label{eq:U(s)}
\end{equation}
where $\gamma \in [0,1]$ is the discount factor that is sometimes used to bias towards earlier rewards, particularly in \textit{infinite horizon} problems ($t \rightarrow \infty$). The optimal utility for a state can then also be computed recursively using
\begin{equation}
U_t^*(s)=\max_{a \in A} (R(s,a)+\gamma \sum_{s'} T(s,a,s')U^*_{t-1}(s')),
\label{eq:U*(s)}
\end{equation}
which for $t \rightarrow \infty$ becomes the \textit{Bellman equation} \cite{Bellman1957}. Knowing the optimal utility function, we can derive an optimal policy:
\begin{equation}
\pi^*(s) = \argmax_{a \in A} (R(s,a)+\gamma \sum_{s'} T(s,a,s')U^*(s')).
\label{eq:pi*(s)}
\end{equation}

In problems with state uncertainty, beliefs $b \in B$ can take the place of states in equations \ref{eq:U(s)}--\ref{eq:pi*(s)} \cite{Kaelbling1998}. The \textit{Markov property} is assumed, meaning that $T(s,a,s')$ does not depend on the sequence of transitions that led to $s$ \cite{Kemeny1983}.

Now that the foundational concepts have been described, we will focus on those most relevant to the proposed unification: states and the reward function $R(s,a)$. States can be vector quantities (Section \ref{sec:systems}). For real-world problems, the relevant elements of the operating environment are sometimes included in the state vector, either explicitly or implicitly \cite{Ragi2014}. For instance, the rover state vector would certainly need to include the rover's $x$ and $y$ coordinates, but may also include time $t$. These three elements allow us to indirectly access other information about the environment, \textit{e.g.}, solar illumination, ambient temperature, communications coverage, and terrain properties.

Similarly, health-related elements can be included in the same state vector. For the rover, the battery charge would likely be in it already for operational purposes. Adding battery temperature, however, would allow for better reasoning about the state of battery health, when combined with information on ambient temperature, terrain, and recharge current. Thus, including even a few health-related elements in the state vector may have a multiplicative effect on the amount of information available. The resulting size of the state vector may also end up being smaller than the sum of separately maintained DM and SHM state vectors, as redundant elements are eliminated.

The reward function $R(s,a)$ encodes the costs and the rewards of being in a particular state or taking a particular action in a state and can also be thought of as a local utility function. For many realistic problems with multi-variable state representations, the reward function needs to combine costs or rewards pertinent to state components. Several approaches have been proposed \cite{Keeney1993}, with additive decomposition being an effective option in many cases. However it is implemented, the key property of the function is that by mapping multiple variables to a single number, it allows us to compute $U(s)$ or $U(s,a)$ and translate a potentially complex DM formulation into an abstract utility maximization problem.

As an example of such mapping, consider two different rover states: $s_1$, where the rate of battery discharge is higher than nominal due to a parasitic load in the electrical system and $s_2$, where the rover is healthy, but is traversing difficult terrain, thus also leading to a higher rate of battery discharge. The utilities $U(s_1)$ and $U(s_2)$ may end up being approximately equal, however, given the similar impact of the two issues on future rewards. Policies computed for $s_1$, $s_2$, and their follow-on states may be similar also and, for instance, could result in more frequent recharge stops.

One notable consequence of health-related components being integrated into a common state vector is that from the computational point of view the concepts of \textbf{fault} and \textbf{failure} become somewhat superfluous. If subsets $S_\text{fault} \subset S$ or $S_\text{failure} \subset S$ are defined for the system, the framework described above will not do anything differently for them. The only essential subset of $S$ is $S_T$ (the terminal states). Failure states may be part of $S_T$ if they result in termination of system operations; however, goal (success states) are members of $S_T$ also. The only difference between them is in their $U(s)$ values. As long as a component fault or a failure does not lead to a transition to a terminal state, actions that maximize the expected value of that state will be selected (which, as it happens, implements the ``fail operational'' philosophy).

\begin{figure*}[htb]
	\centering
		\includegraphics[width=0.7\textwidth]{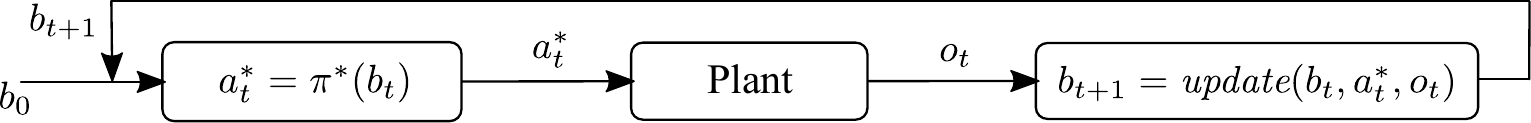}
	\caption{The main loop of health-aware decision making}
	\label{fig:ops_loop}
\end{figure*}

We will refer to this unified approach as \textbf{health-aware decision making (HADM)}. The rest of the major SHM concepts are incorporated into the new approach as follows. \textbf{Fault detection} and \textbf{diagnostics} are subsumed in belief estimation and updating, although these operations are, of course, used for nominal belief states as well. \textbf{Uncertainty management} can now be purposefully incorporated into the decision making process by either augmenting $A$ with information gathering actions \cite{Spaan2015}, evaluated in the same context as other types of actions, or by continuing to improve $U(s)$ or $U(s,a)$ estimates until a desired level of confidence in them is reached \cite{Browne2012}. For actively controlled systems, \textbf{predictive} simulations are simply an integral part of $U(s)$ calculation, where $T(s,a,s')$ serves as a one-step ``prognostic'' function (with degradation models, if any). Whereas prognostic algorithms applied to controlled systems are limited in their predictive ability due to the lack of knowledge about future actions, here the $U(s)$ calculation process is an exploration of possible execution scenarios, thus combining $s'$ or $b'$ estimation with sequential action selection.

The overall HADM operational loop (assuming state and outcome uncertainty) can be seen in Figure \ref{fig:ops_loop}. Once the initial belief $b_0$ is estimated at time $t_0$, either an offline policy is referenced or an online policy is computed to determine $a_0^*$ (best action). The action is executed by the plant, transitioning to a new (hidden) state, and generating an observation $o_0$. The observation is then used to update $b_0$ (typically through some form of Bayesian updating) and the process repeats until a terminal state is believed to be reached.

A detailed discussion of the actual algorithms that can implement the proposed approach is left for future publications, however examples of DM algorithms that can deal with state or outcome uncertainty are provided by \citeauthor{Kochenderfer2015} \shortcite{Kochenderfer2015}. In systems where both state and outcome uncertainty are not a factor, a variety of traditional state space planning algorithms \cite{Ghallab2016} would not need any modification to produce plans for spaces of state vectors that include health-related components.

\subsection{Emergency response vs. health management}

For realistic complex systems operating in the presence of state and outcome uncertainty, $S$, $B$, and $O$ are likely to be infinitely large (with $|A| \ll \infty$, although still potentially large). The problem of finding exact optimal policies in such cases is PSPACE-complete \cite{Papadimitriou1987}. Approximate solution methods typically work \textit{online}, constructing $\pi$ for the current belief $b$ based on beliefs reachable from $b$ within the decision horizon \cite{Browne2012}. They also typically perform targeted sampling from $S/B$, $A$, and $O$, thus optimality guarantees can be harder to provide. We, therefore, propose the following:
\begin{itemize}
	\item That \textbf{system emergency response (SER)} be defined as an automated or semi-automated process that is invoked to maximize the likelihood of preserving the system's integrity, regardless of the effect on operational goals (\textit{i.e.}, a different objective from HADM).
	\item That emergency response policy $\pi_\text{SER}$ be computed separately from the HADM policy.
\end{itemize}
In the space domain, an example of SER is putting a spacecraft in a \textit{safe mode} until the emergency is resolved \cite{Rasmussen2008}. In aviation, it could be executing recommendations of a collision avoidance system \cite{Kochenderfer2015}. Introduction of a separate SER system would likely require the introduction of $S_\text{SER}$, an additional $S$ subset which defines for which states $\pi_\text{SER}$ is invoked versus $\pi_\text{HADM}$. Once again, however, $S_\text{SER}$ will not necessarily only contain system fault and failure states. For instance, states where environmental conditions warrant emergency response (\textit{e.g.}, solar activity interrupting teleoperation of the rover) would be included as well. Since the scope of the SER problem is likely to be much narrower than that of HADM, it opens up the possibility of computing, verifying, and validating $\pi_\text{SER}$ \textit{offline} \cite{Kochenderfer2015}.

As sensors, computing capabilities, and DM algorithms improve further, the fraction of the system's state space that is under the purview of SER will decrease. Still, we foresee the need for a ``safety net'' SER to be there for safety-critical functions and thus advocate SER independence. SER would cover the situations where the primary HADM system may not be able to produce a suitable solution in a desired amount of time, serving as an equivalent of human reflexive responses triggered by important stimuli versus the more deliberative cognitive functionality of the brain. It could also provide dissimilar redundancy for critical regions of $S$, essentially implementing the Swiss Cheese Model \cite{Reason1990a}, which advocates multiple different layers of protection for important functions.

\subsection{Remarks on computational complexity}

A counter-argument to the unified approach can be made that it increases the computational complexity by operating on larger $S/B$, $A$, and $O$ spaces. While this is indeed a concern, algorithms have been developed in the recent years that can effectively accommodate much of the complexity, even for problems with state and outcome uncertainty, by approximating optimal policies online \cite{Silver2010,Browne2012}. Secondly, some of the complexity is removed from HADM into an independent SER component, making the former problem easier. Finally, as noted above, unifying the SHM and DM elements may also result in reduction of overlapping model variables.

\section{Conclusions}
\label{sec:conclusions}

This paper reexamines the prevalent approach to performing system health management and makes the case for why keeping system health management functions separate from decision making functions can be inefficient and/or ineffective. We also present the case for why prognostics is only meaningful in a limited set of circumstances and, even then, needs to be driven by decision making requirements.

We then explain the rationale for unifying (not just integrating) system health management with decision making and outline an approach for accomplishing that. We also propose keeping emergency response functionality separate to guarantee timely response, provide dissimilar redundancy, and allow for offline computation, validation, and verification of emergency response policies.

We believe that the proposed unification approach will improve the performance of both decision making and system health management, while potentially simplifying informational architectures, reducing sensor suite sizes, and combining modeling efforts. The approach is scalable with respect to system complexity and the types of uncertainties present. It also puts a variety of existing and emerging computational methods at the disposal of system designers.

\clearpage
\fontsize{9.0pt}{10.0pt} \selectfont
\bibliography{../../bibliography/library_fixed}
\bibliographystyle{aaai}

\end{document}